\pdfoutput=1

\documentclass[11pt,a4paper,usenames,dvipsnames]{article}
\usepackage[hyperref]{emnlp-ijcnlp-2019}
\usepackage{times}
\usepackage[T1]{fontenc}
\usepackage[utf8]{inputenc}
\usepackage{latexsym}
\usepackage{linguex}
\usepackage{enumitem}
\usepackage{booktabs}
\usepackage{graphicx}

\usepackage{url}
\usepackage{xcolor}
\usepackage{tikz,pgfplots}
\usepackage{tikz-dependency}
\usepackage{tikz-qtree}
\usepackage{multirow}
\usepackage{subcaption}
\usepackage{multicol}
\usepackage{floatrow}
\usepackage{array,booktabs,calc}
\usepackage{gb4e}
\noautomath \usepackage{cgloss4e}

\definecolor{imsblue1}{RGB}{43,52,126}
\definecolor{imsblue2}{RGB}{41,79,157}
\definecolor{imsblue3}{RGB}{43,119,188}
\definecolor{imsblue4}{RGB}{0,154,204}
\definecolor{sidebarblue}{RGB}{23,84,165}

\definecolor{imsred1}{RGB}{178,32,31}
\definecolor{imsred2}{RGB}{197,46,35}
\definecolor{imsred3}{RGB}{209,10,17}
\definecolor{imsred4}{RGB}{226,6,19}

\definecolor{mygreen}{RGB}{50,180,50}
\definecolor{myseagreen}{RGB}{046,139,087}
\definecolor{myforest}{RGB}{34,139,34}
\definecolor{seablue}{RGB}{65,105,225}
\definecolor{mygray}{gray}{0.9}

\aclfinalcopy

\title{Pragmatic information in translation: \\ a corpus-based study of tense and mood in English and German}

\author{Anita Ramm\textsuperscript{\dag}, Ekaterina Lapshinova-Koltunski\textsuperscript{\ddag}, Alexander Fraser\textsuperscript{\dag}\\
	\textsuperscript{\dag}Center for Information and Language Processing, LMU Munich \\
	\textsuperscript{\ddag}Saarland University  \\
}

\date{}

\begin{document}
\maketitle
\begin{abstract}
  Grammatical tense and mood are important linguistic phenomena to
  consider in natural language processing (NLP) research. We consider the
  correspondence between English and German tense and mood in
  translation.  Human translators do not find this correspondence
  easy, and as we will show through careful analysis, there are no
  simplistic ways to map tense and mood from one language to
  another. 
  Our observations about the challenges of human translation
  of tense and mood have important implications for multilingual
  NLP. Of particular importance is the
  challenge of modeling tense and mood in rule-based, phrase-based 
  statistical  and neural machine translation.
\end{abstract}

\section{Introduction}
\label{intro}

This paper analyzes tense and mood in English and German from the perspective of the data commonly used to train
MT systems or to model tense/mood,
namely freely available bilingual texts. The need for a thorough analysis of tense/mood in
parallel texts
arises
from the fact that there is a high degree of variation between the two languages resulting in a many-to-many relation in the tense/mood translation between English and German. Particularly, frequently occurring unintuitive tense correspondences and 
the
low frequency of the
many tense/mood combinations
is
problematic for different NLP tasks using parallel corpora. We
study the 
correspondences in a large English-German parallel corpus and explain them from the point of view of different pragmatic factors -- contextual constraints in terms of genre/user preferences or textual properties, and tense interchangeability. We compare English and German morpho-syntactic tense sets suffering from tense correspondence gaps in both directions and discuss the impact of translation process on the tense/mood variability in
our data. Finally, we take a look to the modeling of tense and mood for machine translation pointing to important features needed to transfer tenses between languages. Our analysis indicates that  bilingual modeling of tense and mood 
cannot be properly done by considering solely lexical/syntactic features, e.g. words, POS tags, etc.,  
also supported by the previous work~\cite{ye06:sighan}. 
Instead, incorporation of
pragmatic information
is required, which is currently
not directly accessible to
most
NLP systems. We summarize the
pragmatic  information
required
and provide a list of
available tools for 
automatic
annotation 
with the respective information,
which will be of direct use in future efforts to solve this difficult modeling task. In the following, we present theoretical issues and related work (Section \ref{sec:theory}), quantitative analysis on the usage of the tense/mood correspondences in English-German parallel data and their modeling in the context of MT (Section \ref{sec:analyses}), summarizing the findings in Section \ref{sec:conclusion}.

\section{Theoretical issues and related work}
\label{sec:theory}

\subsection{Contrasts in English and German tense and mood systems}
\label{sec:contrastsENDE}

As known from contrastive grammars~\cite{koenig12:endeconstr,hawkins2015comparative}, English and German share a common ground of six morpho-syntactic tenses: present/\emph{Präsens}, simple past/\emph{Präteritum}, present perfect/\emph{Perfekt}, past perfect/\emph{Plusquamperfekt}, future I/\emph{Futur I} and future II/\emph{Futur II}. We summarize those 
with examples in both languages in Table \ref{tab:EnDeTenses}, 
which 
we 
created to show the correspondence between these languages. 
\begin{table*}[!ht]
	\centering
	\tiny
	\begin{tabular}{|m{1cm}||m{3.8cm}|m{3.5cm}||m{1.7cm}|m{3.3cm}|}
		\hline
		\multicolumn{1}{|c||}{\bf Morph.} & \multicolumn{2}{c||}{\bf English} & \multicolumn{2}{c|}{\bf German} \\
		\cline{2-5}
		\multicolumn{1}{|c||}{\bf tense} & \textbf{Synt. tense} & 
		\multicolumn{1}{c||}{\bf Example} & 
		\textbf{Synt. tense} & \multicolumn{1}{c|}{\bf Example} \\
		\hline \hline
		& present simple & (I) read & Präsens & (Ich) lese \\
		\cline{2-5}
		& present progressive & (I) am reading & &  \\
		\cline{2-5}
		& present perfect & (I) have read & Perfekt & (Ich) habe gelesen  \\
		\cline{2-5}
		& present perfect progressive & (I) have been reading & &  \\
		\cline{2-5}
		present & future I & (I) will read \newline (I) am going to read & Futur I & (Ich) werde lesen \\
		\cline{2-5}
		& future I  progressive & (I) will be reading
		\newline  (I) am going to be reading & & \\
		\cline{2-5}
		& future II & (I) will have read & Futur II & (Ich) werde gelesen haben \\
		\cline{2-5}
		& future II  progressive & (I) will have been reading & & \\
		\hline
		& past simple & (I) read & Präteritum & (Ich) las \\
		\cline{2-5}
		& past progressive & (I) was reading & & \\
		\cline{2-5}
		past & past perfect & (I) had read & Plusquam- \newline perfekt & 
		(Ich) hatte gelesen \\
		\cline{2-5}
		& past perfect progressive & (I) had been reading & & \\
		\hline \hline
		present* & conditional I & (I) would read & Konjunktiv II 
		& (Ich) würde lesen \\
		\hline
		& conditional I progressive & (I) would be reading & & \\
		\cline{2-5}
		past* & conditional II & (I) would have read & Konjunktiv II 
		& (Ich) hätte gelesen \\
		\cline{2-5}
		& conditional II progressive & (I) would have been reading & & \\
		\hline
		\multirow{1}{*}{present*} & & & \multirow{1}{*}{Konjunktiv I} & (Er) lese \newline (Er) werde lesen \\  
		\hline
	\end{tabular}
	\caption{List of the tenses in English and German in active voice. The
		table indicates the tense correspondences in terms of their morpho-syntactic structure.}
	\label{tab:EnDeTenses}
\end{table*}
In English, each of the tenses has a progressive variant. The German tense system does not have an explicit marking of the progressive aspect. But German has a larger set of subjunctive tense forms. While a few of them have direct morpho-syntactic counterparts in English, most of them correspond to indicative tenses in English. 
The meaning of a specific tense form may considerably vary too. We summarize the contrasts related to the meaning of the English and German tenses described by \newcite{koenig12:endeconstr} in Table \ref{tab:TMuseENDE}. This description refers to different aspects such as the time reference (\emph{past, futurate, future}, etc.) and relation to the moment of utterance (\emph{resultative, universal, narrative}). In other words, the (non-)parallelism of the 
respective 
tenses can be established 
by considering specific semantic properties of a given verb and the utterance that the respective verb occurs in. Different aspects in the English tense system have
different impacts
on
the use of tenses. For instance, in contrast to the simple present tense, the present progressive can be used in the futurate context. In German, \emph{Präsens} can almost always be used to refer to
the
future. The English progressive tense lacks
direct counterparts in German and is therefore translated into a number of different German tenses. 

English and German also differ greatly with respect to the grammatical mood. In German, the subjunctive is expressed in the verbal morphology and interacts with the German tense system changing the time of an utterance. German distinguishes between two subjunctive morpho-syntactic forms: \emph{Konjunktiv I} and \emph{Konjunktiv II}. The latter is used in the context of
conditional and contrafactual utterances. 
Usually, sentences with \emph{Konjunktiv II} are composed of at least two clauses. There are, however, also \emph{free factive} occurrences of \emph{Konjunktiv II}, where it occurs in a simple sentence, see Example (\ref{konjIIfreefact}). Such sentences may, for instance, indicate politeness. 

\begin{table*}[!ht]
	\tiny
	\centering
	\begin{tabular}{|m{2cm}||m{6.2cm}|m{5.6cm}|}
		\hline
		\multicolumn{1}{|c||}{\bf Use} & \multicolumn{1}{c|}{\bf  German} & 
		\multicolumn{1}{c|}{\bf English} \\
		\hline \hline
		\multicolumn{3}{|l|}{\bf Präsens/present tense}\\
		\hline
		non-past & Ich schlafe von 12 bis 7. & I sleep from midnight to seven. \\
		futurate & Morgen weiß ich das. & $\rightarrow$ future tense \tiny{(\emph{I will know that tomorrow.})}\\
		\hline
		\multicolumn{3}{|l|}{\bf Präteritum/simple past }\\
		\hline
		past time & Ich schlief den ganzen Tag. & I slept the whole day. \\
		\hline
		\multicolumn{3}{|l|}{\bf Futur I/future tense} \\
		\hline
		future time & Ich werde schlafen. & I will sleep. I am going to sleep.\\
		\hline
		\multicolumn{3}{|l|}{\bf Perfekt/present perfect} \\
		\hline
		resultative & Jemand hat mein Auto gestohlen. & Someone has stolen my car.\\
		existential & Ich habe (schon mal) Tennis gespielt. & I have played tennis. \\
		hot news & Kanzler Schröder ist zurückgetreten. & Chancellor Schröder has resigned. \\
		universal & $\rightarrow$ Präsens \tiny{(\emph{Ich lebe hier seit 2 jahren.})} & I have lived here for two years.\\
		narrative & Ich bin gestern im Theater gewesen. & $\rightarrow$ past tense \tiny{(\emph{I was in theater yesterday.})} \\
		\hline
		\multicolumn{3}{|l|}{\bf Futur II/future perfect} \\
		\hline
		future \newline results & Ich werde das bis morgen erledigt haben. & I will have done this by tomorrow. \\
		\hline
		\multicolumn{3}{|l|}{\bf Plusquamperfekt/past perfect} \\
		\hline
		pre-past & Ich hatte geschlafen. & I had slept.\\
		\hline
	\end{tabular}
	\caption{\label{tab:TMuseENDE} Meaning of tenses in English and German \protect\cite[p. 92]{koenig12:endeconstr}}
\end{table*}

\begin{exe} 
					\ex{
				\gll Ich \textbf{hätte} gern ein Glas Wasser.
				                \\
				I have gladly a glass water. \\
                \glt 'I'd like to have a glass of water.'}
				\label{konjIIfreefact}
								\end{exe}
		
\noindent
Both \emph{Konjunktiv I} and \emph{Konjunktiv II} can be used in the context of the reported speech. Note, however, that the use of the subjunctive mood is not grammatically required to signal reported speech. In fact, the two \emph{Konjunktiv} forms and the indicative mood  are often used interchangeably in reported speech \cite{csipak15:gram}. 						
For
the English
subjunctive mood,~\newcite{koenig12:endeconstr} rather use the term \emph{quasi-subjunctive}, since subjunctive mood in English exists only for the verb \emph{be}. Other forms used in the subjunctive contexts correspond to the infinitives. The German \emph{Präsens} and \emph{Futur I} are interchangeable in many contexts. In the futurate use, \emph{Präsens} is usually combined with a temporal phrase which points to the future; in (\ref{dePresFut1a}), the adverbial \emph{morgen} provides the respective temporal information. However, the temporal phrase is not always overtly given in a considered sentence: in (\ref{dePresFut1b}), the verb \emph{kommen} in the present tense refers to the future which is obvious solely by considering the preceding sentence. 

\begin{exe} 
	\ex{
				Ich \textbf{komme} morgen. 
         				I come tomorrow. 								'I'll come tomorrow.'
				\label{dePresFut1a}
				}
\end{exe}
\begin{exe} 
	\ex{
		\gll \textbf{Kommst} du morgen? 
					Ja, ich \bf{komme}. 
                                        \\
					Come you tomorrow? Yes, I come.\\
					\glt 'I'll come tomorrow.'
					\label{dePresFut1b}
					}
\end{exe}

\noindent
Another prominent example of tense interchangeability in German is related to the past tenses. 
There are some fine-grained differences between the respective tenses, but at least \emph{Präteritum} and \emph{Perfekt} are interchangeable in many contexts, see~\newcite{sammon02:engram}. In fact, the dominance of either of the forms is a matter of author's preference or contextual constraints, see \ref{sec:register} below. For instance, \emph{Perfekt} is often used in spoken language, while \emph{Präteritum} is more frequently used in writing. Furthermore, there is a certain lexical preference: auxiliaries and modals are more frequently used in \emph{Präteritum} than in \emph{Perfekt}.

\subsection{Contextual constraints}
\label{sec:register}

Contextual constraints on the tense/mood usage have been analyzed mostly in a monolingual context. For example, \newcite{weinrich64:tempus} differentiates between two groups of the German tenses: (i) \emph{discussing} (\emph{Präsens},
\emph{Perfekt}, \emph{Futur I}, \emph{Futur II}) and (ii) \emph{narrative} (\emph{Präteritum}, \emph{Plusquamperfekt}, and subjunctives \emph{Konjunktiv I} and \emph{Konjunktiv II}). His classification is relevant for genre differentiation. 
For instance, the narrative tenses are mostly found in written German (e.g., literary works), while the discussing tenses are more often used in the spoken language.  In a multilingual context, there exist a few studies that analyze the role of tense/mood in functional variation of language called register variation. ~\newcite{biber1995} uses preferences for specific tense and mood as linguistic indicators for specific registers in a number of languages.  \newcite{neumann13:regVar} presents a contrastive corpus-based study of English and German (including translations), in which the tense frequency is used among other textual properties  to induce the \emph{goal type} of the text (one of the parameters of register variation): argumentation, narration, instruction, etc. She observed that the frequency of present vs.  past across texts from different registers expose different (i.e., domain-specific) distributional specifics: past tenses are rather typical for narrative texts, while present tense verbs are more typical for argumentative texts such as political essays, popular science articles, etc. These findings are in line with the classification of tenses proposed by \newcite{weinrich64:tempus}. In addition to contextual constraints expressed in genre or register, translation of tenses may also follow a \emph{set of rules} defined for a specific translation project. For instance, the translation guidelines of the European Commission for German require
the session minutes or reports be written in the present tense.\footnote{\url{https://ec.europa.eu/info/sites/info/files/german_style_guide_de_0.pdf}}

\subsection{Tense and mood in human translation}
\label{sec:transCharacteristicsLing}
Tense and mood were analyzed in
previous
studies on English-German translation~\cite{teich03,neumann13:regVar}. However, a systematic description of tense/mood transformation patterns for this language pair
has been missing until our work.
At the same time, translation studies provide us with valuable information on how translation process has an impact onto translated texts which, as a result, differ from non-translated texts both in the source (SL) and the target language (TL). These differences are reflected in the features of translated language~\cite{gellerstam86,baker93}. Two of these translation features are important for bilingual modeling of tense and mood: (i) \emph{shining through} and (ii) \emph{normalization}. The former one indicates the closeness of the translation to the source~\cite{teich03}, whereas the latter one is related to the tendency to conform (and exaggerate) the patterns typical for the TL~\cite{baker93}. 
We would observe \emph{shining through} in our data if tenses used in the sources are preserved in the translations.  While there is much parallelism with respect to tense in the two languages under analysis, many cases may expose a TL-specific usage of tense, which may considerably differ from a form given in the source due to a smaller set of tenses available in the German language system. Following~\newcite{teich03}, \emph{shining through} is less prominent and \emph{normalization} is more prominent when translating into a language which has fewer options with respect to a specific grammatical system. 
This means that our parallel texts may expose a great variation in the tense translation. 
 Finally, parallel corpora represent a concatenation of the translations produced by many different translators. Therefore, we expect that the observed variation in tense translations can be impacted by the preferences of a specific translator.
\subsection{Tense and mood in machine translation and NLP}
\label{sec:MTandNLP}

In the context of the rule-base MT, i.e., in EUROTRA~\cite{copeland-et-al:1991}, translation of tense and mood relies on an interlingua representation to which the SL sentence is mapped, and which is then mapped to the syntax of the TL respectively.  This mapping is rule-based and follows a set of manually defined rules which make use of different kinds of information. The rules for English-German formulated within EUROTRA indicate that tense cannot be considered in isolation, but rather in a combination with other related linguistic features such as aspect and {\sl Aktionsart}. Thus, specific modality, as well as voice properties, need to be considered in the bilingual modeling of tense and mood.

Recently, there have been attempts to automatically model tense and mood for different NLP tasks. In the monolingual context, for instance, \newcite{tajiri12:acl} used a tense classification model  for detecting and correcting tense in the texts produced by English learners. In the bilingual context, \newcite{ye06:sighan} presented an empirical study of the features needed to train a classification model for predicting English tenses given the source sentences in Chinese. \newcite{degispertmarino:08}, \newcite{loaiciga14:lrec} and \newcite{ramm16:wmt} presented work on building tense classification models which are used to improve tense choice in statistical MT systems for English-Spanish, English-French and English-German, respectively. While \newcite{loaiciga14:lrec} reported
encouraging results, \newcite{degispertmarino:08} and \newcite{ramm16:wmt} left unanswered questions
about the appropriate method and the
necessary
contextual information for modeling tense and mood in a bilingual context.

\section{Analyses}\label{sec:analyses}
\subsection{Tense and mood in human translation}
\label{sec:tmHT}
\paragraph*{Data and tools}
\label{sec:corpTools}

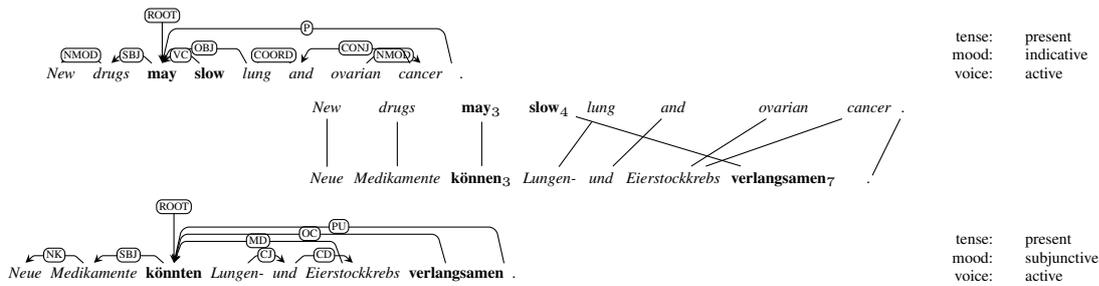
\begin{figure*}[!ht]
\centering
\begin{minipage}{0.7\textwidth}
	
	\begin{dependency}
			\tiny
			\hspace{0.5cm}
			\begin{deptext}[column sep=4,row sep=1,font=\it]
				New \& drugs \& {\bf may} \& {\bf slow} 
				\& lung \& and \& ovarian \& cancer  \& . \\
											\end{deptext}
			\deproot[edge unit distance=2ex]{3}{ROOT} 
			\depedge[edge unit distance=1ex]{3}{2}{SBJ}
			\depedge[edge unit distance=1ex]{2}{1}{NMOD}
			\depedge[edge unit distance=1ex]{6}{5}{COORD}
			\depedge[edge unit distance=1ex]{5}{3}{OBJ}
			\depedge[edge unit distance=1ex]{4}{3}{VC}
			\depedge[edge unit distance=1ex]{7}{8}{NMOD}
			\depedge[edge unit distance=1ex]{8}{6}{CONJ}
			\depedge[edge unit distance=0.8ex]{9}{3}{P}
			
		\end{dependency}

\end{minipage}
\hfill
\begin{minipage}{0.23\textwidth}
\vspace{0.2cm}
\tiny
	\begin{tabular}{rl}
        tense: &present \\
        mood: & indicative \\
        voice: & active \\
        \end{tabular}
\end{minipage}

\begin{tikzpicture} 
    \tiny
	\matrix(m)[matrix of nodes,row sep=0.6cm, column sep=0.001ex,ampersand replacement=\&,font=\it]
	{
		New \& drugs \& {\bf may$_3$} \& {\bf slow$_4$} \& lung \& and \& ovarian \& cancer \& . \\
		Neue \& Medikamente \& {\bf können$_3$} \& Lungen- \& und \& Eierstockkrebs \& {\bf verlangsamen$_7$} \& . 
		\\
	};
	\draw (m-1-1) -- (m-2-1); 	\draw (m-1-2) -- (m-2-2); 	\draw (m-1-3) -- (m-2-3); 	\draw (m-1-4) -- (m-2-7); 	\draw (m-1-5) -- (m-2-4); 	\draw (m-1-6) -- (m-2-5); 	\draw (m-1-7) -- (m-2-6); 	\draw (m-1-8) -- (m-2-6); 	\draw (m-1-9) -- (m-2-8); 	\end{tikzpicture}
	
\begin{minipage}{0.7\textwidth}
		\begin{dependency}
			\tiny
			\begin{deptext}[column sep=0.6,row sep=1,font=\it]
				Neue \& Medikamente \& {\bf könnten}
				\& Lungen- \& und \& Eierstockkrebs \& 
				{\bf verlangsamen} \& . \\
																							\end{deptext}
			\deproot[edge unit distance=2.3ex]{3}{ROOT} 
			\depedge[edge unit distance=1ex]{3}{2}{SBJ}
			\depedge[edge unit distance=1ex]{2}{1}{NK}
			\depedge[edge unit distance=1ex]{6}{3}{MD}
			\depedge[edge unit distance=1ex]{5}{6}{CD}
			\depedge[edge unit distance=1ex]{4}{5}{CJ}
			\depedge[edge unit distance=1ex]{7}{3}{OC}
			\depedge[edge unit distance=1ex]{8}{3}{PU}
			
		\end{dependency} 
\end{minipage}
\hfill
\begin{minipage}{0.23\textwidth}
\vspace{0.4cm}
\tiny
	\begin{tabular}{rl}
        tense: & present \\
        mood: & subjunctive \\
        voice: & active \\
        \end{tabular}
\end{minipage}
		
\caption{Word-aligned, parsed English-German parallel sentence pair with TMV annotations. 
Parallel VC: \emph{may slow $\leftrightarrow$ können verlangsamen}, tense/mood pair \emph{present/indicative $\leftrightarrow$ present/subjunctive}.}
\label{fig:parallelSents}
\end{figure*}

Since one of our aims is to serve the task of machine translation, 
our contrastive analysis of tense and mood in English and German relies on the parallel corpora provided for WMT15 shared tasks on machine translation~\cite{wmt15}.
We make use of the News corpus (news articles, 272k sentences), the Europarl corpus (1,9 mio. sentences) and the Crawl corpus, a large collection of mix-domain bilingual documents retrieved from the Internet (2,4 mio. sentences). In addition, we also consider Pattr\footnote{\url{http://www.cl.uni-heidelberg.de/statnlpgroup/pattr/de-en.tar.gz}}, a medical corpus (1,8 mio. sentences). In this way, we have a constellation of various domains (as they are called in NLP) or registers/genres (as they are called in the studies described in \ref{sec:register} above).
The corpora are tokenized with a standard tokenizer provided with the SMT toolkit \emph{Moses}
\cite{koehn07:acl} and parsed with the Mate parser
\cite{bohnet12:emnlp} which provides dependency parse trees for both languages,
and, for German, 
morphological analysis of
words. Both sides of the parallel corpora are annotated with tense, mood and voice information using the TMV annotator
\cite{ramm17:acl}. The English-German verb pairs annotated with the respective information are then extracted by (i) automatically computing word alignment of the parallel texts with Giza++
\cite{och03:asc} and (ii) identifying pairs of
VCs from the aligned, annotated parallel data
(see example in 
Figure \ref{fig:parallelSents}). In our analyses,
we do not differentiate between translation directions,
because
we are interested in all transformations possible for the analyzed language pair.

\paragraph*{Indicative tense}
\label{sec:lingAspects}

\begin{figure*}[!ht]
	\centering
	\begin{tikzpicture}
    \tiny
	\begin{axis}[
	ybar=.1cm,ymode = log,log origin=infty,ymajorgrids,
	bar width=.4cm,
	width=\textwidth,
	height=.45\textwidth,
	legend style={at={(0.88,1)},
		anchor=north,legend columns=-1},
	symbolic x coords={Präsens, Präteritum, Perfekt, Pluperfekt,Futur I, Futur II, Konj I pres, Konj I past, Konj II pres, Konj II past},
	xtick=data,
	ytick={0.001,0.005,0.01,0.05,0.1,0.3,0.6},
	yticklabels={0.001,0.005,0.01,0.05,0.1,0.3,0.6},
	ymin=0,ymax=1,
	label style={yshift=-0.5em},xticklabel style={rotate=0}
	]
	\addplot[imsred1,fill=imsred1!40!white,postaction={pattern=north west lines}] table[x=Tense,y=PresPerf]{\mydataPresPerf};
	\addplot[imsblue1,fill=imsblue1!40!white,postaction={pattern=north east lines}] table[x=Tense,y=PresPerfProg]{\mydataPresPerf};
	\legend{PresPerf, PresPerfProg}
	\end{axis}
	\end{tikzpicture}
	\caption{				        German correspondences of the English present prefect 
        (progressive) in the Europarl corpus.}
	\label{fig:presPerfEP}
\end{figure*}
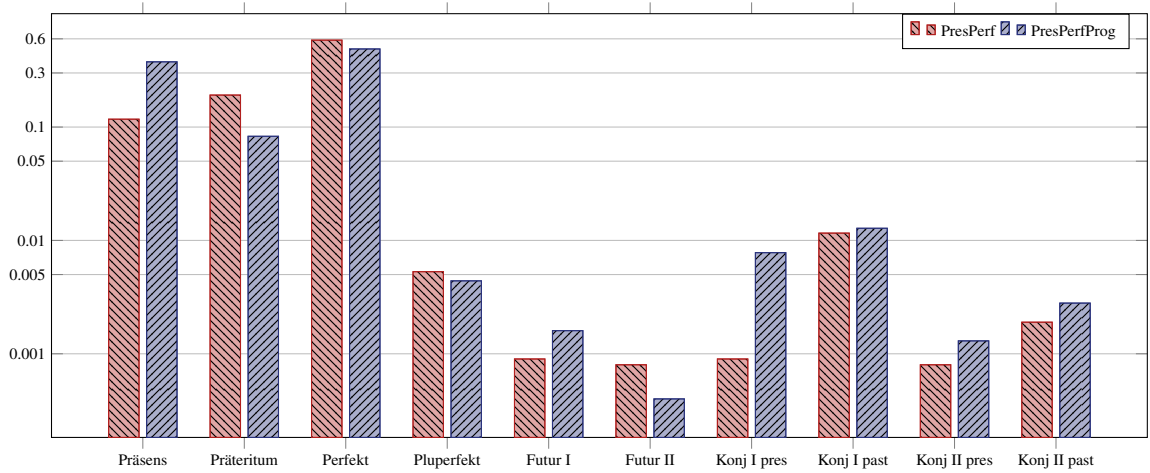

As already mentioned in Section
\ref{sec:contrastsENDE},
the English progressive tenses are translated into a number of different German tenses. 
Figure \ref{fig:presPerfEP} illustrates the frequency distribution of the English present perfect (progressive) in our data. 
It
is striking
that both English tense forms correspond to three different German tenses in most cases: \emph{Präsens}, \emph{Perfekt} and \emph{Präteritum}
whereby 
\emph{Perfekt} is the most prominent equivalent. Considering the two German past tenses together, it becomes clear that both present perfect tense forms correspond to one of the German past tenses more often than the present tense does. Progressiveness also seems to have a large impact on the translation of present perfect into German: ca.\  77\% of the non-progressive forms corresponds to one of the German past tenses, whereas 56\% of the progressive cases do so. In other words, the progressive present tense still prefers to be transferred into one of the German past tenses. However, the German \emph{Präsens} more often corresponds to this English tense than to the non-progressive variant. 

\pgfplotstableread[row sep=\\,col sep=&]{
	Tense & PresPerf & PresPerfProg \\
	Präsens &  0.11731 & 0.3762 \\
	Präteritum & 0.1915 & 0.0829 \\
	Perfekt & 0.5842 & 0.4885 \\
	Pluperfekt & 0.0053 & 0.0044 \\
	Futur I & 0.0009 & 0.0016 \\
	Futur II & 0.0008 & 0.0004 \\
	Konj I pres & 0.0009 & 0.0078 \\
	Konj I past & 0.0116 & 0.0128 \\
	Konj II pres & 0.0008 & 0.0013 \\
	Konj II past & 0.0019 & 0.0028 \\
}\mydataPresPerf

\paragraph*{Subjunctive mood}
\label{sec:subjTenses}

\begin{figure*}[!ht]
	\centering
	\begin{tikzpicture}
    \tiny
	\begin{axis}[
	ybar stacked,
	bar width=.4cm,
	width=\textwidth,
	height=.45\textwidth,
	legend style={at={(0.93,1)},
		anchor=north,legend columns=-1,font=\tiny},
	symbolic x coords={pres, presProg, past, pastProg, presPerf, presPerfProg, pastPerf, pastPerfProg, futureI, futureIProg, futureII, futureIIProg, condI, condIProg, condII, condIIProg, gerund,  toInfinitive},
	xtick=data,legend columns=1, legend cell align=left,
	ymin=0,ymax=1,
	xticklabel style={rotate=40}
	]
	\addplot[imsblue1,fill=imsblue1!40!white,postaction={pattern=north west lines}] table[x=Tense,y=Konjunktiv I]{\mydataConj};
	\addplot[imsred1,fill=imsred1!40!white,postaction={pattern=north east lines}] table[x=Tense,y=Konjunktiv II]{\mydataConj};
	\legend{Konjunktiv I, Konjunktiv II}
	\end{axis}
	\end{tikzpicture}
	\caption{		        English correspondences of the German \emph{Konjunktiv} 
        tenses in the Europarl corpus.}
	\label{fig:konjDistribNews}

\end{figure*}
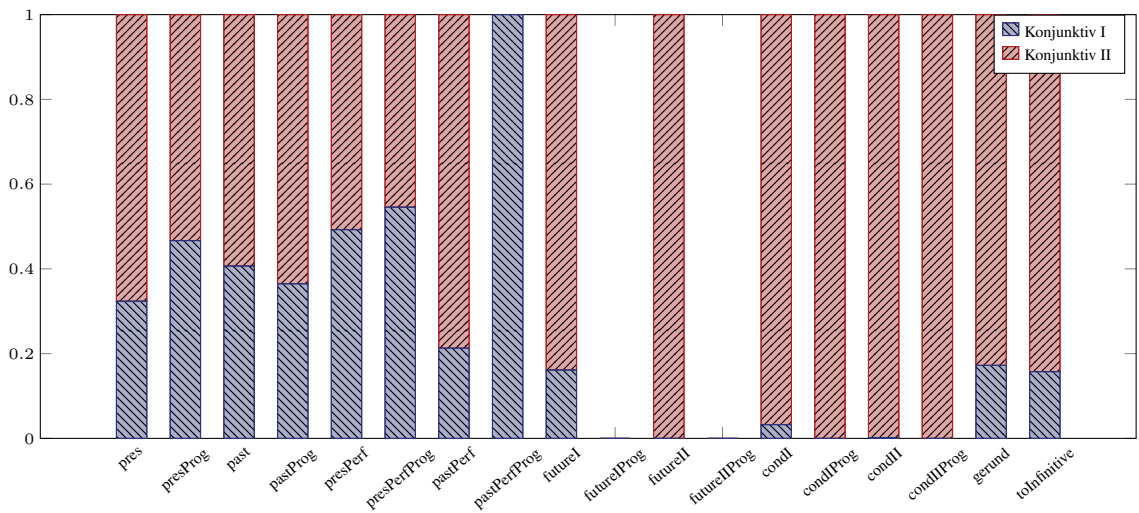

The frequency distribution of the tense correspondences between subjunctive forms based on news texts is shown in Figure \ref{fig:konjDistribNews}. As expected, 
the German \emph{Konjunktiv} tense forms are equivalents of 
all English indicative tense forms 
in the dataset at hand. Thereby, the  \emph{Konjunktiv II} is a more frequent equivalent than the \emph{Konjunktiv I}. Assumed that the conditional and counter-factual situations in English are described with conditional forms, it is quite unexpected that the other English tense forms more often  correspond to the German \emph{Konjunktiv II} (used to indicate conditional contexts) than to \emph{Konjunktiv I} (used to indicate reported speech) in our translation data. A possible explanation for this is that in the news data, \emph{Konjunktiv II} is more often used to express reported speech than the \emph{Konjunktiv I} form. When expressing non-factual events, English conditionals can be seen as direct counterparts of the German \emph{Konjunktiv II}: Figure  \ref{fig:condDistribNews} shows that \emph{Konjunktiv II} is the most frequent equivalent for all four English conditional tense forms in our data. Further frequent counterparts are \emph{Präteritum} for the conditional I and \emph{Perfekt} for the  conditional II.

\pgfplotstableread[row sep=\\,col sep=&]{
	Tense & Konjunktiv I & Konjunktiv II \\
	pres & 0.323552324072 & 0.676447675928  \\
	presProg & 0.466666666667 & 0.533333333333  \\
	presPerf & 0.492481203008 & 0.507518796992 \\
	presPerfProg & 0.545454545455 & 0.454545454545 \\
	past & 0.406629834254 & 0.593370165746 \\
	pastProg & 0.364705882353 & 0.635294117647  \\
	pastPerf & 0.212719298246 & 0.787280701754  \\
	pastPerfProg & 1.0 & 0  \\
	futureI & 0.161234991424 & 0.838765008576 \\
	futureIProg & 0 & 0  \\
	futureII & 0 & 1.0 \\
	futureIIProg & 0 & 0  \\
	condI & 0.032373785983 & 0.967626214017  \\
	condIProg & 0 & 1.0 \\
	condII & 0.00145348837209 & 0.998546511628  \\
	condIIProg & 0 & 1.0  \\
	gerund & 0.172185430464 & 0.827814569536 \\
	toInfinitive & 0.157428970657 & 0.842571029343  \\	
}\mydataConj

\paragraph*{Finite vs. non-finite verbal complexes}
\label{sec:infVCs}

Our data shows that the usage of non-finite VCs in the two languages varies considerably.
For instance, in the News corpus, 16.7\% of all VCs in English are non-finite, while this is the case for only 7.9\% of the German VCs. Similar ratio 
is also given in the Europarl corpus in which 18.2\% of the English VCs and 6.2\% of the German VCs, respectively, are non-finite.
Figure \ref{fig:nonfinNewsEP} indicates that the major part of the English non-finite VCs have German finite VCs as equivalents. These translation equivalents pose an interesting problem in the context of MT. When translating from English to German, MT needs to generate a finite clause for the given non-finite source clause. Particularly, it needs to generate a finite German VC in a tense form for which there is no obvious evidence in the source. 

\pgfplotstableread[row sep=\\,col sep=&]{
	Tense & News & Europarl \\
	Präsens &  0.3070 & 0.5060 \\
	Präteritum & 0.1019 & 0.0569 \\
	Perfekt & 0.0365 & 0.0030 \\
	Pluperfekt & 0.0069 & 0.0026 \\
	Futur I & 0.0408 & 0.0410 \\
	Futur II & 0.0025 & 0.0022 \\
	Konjunktiv I & 0.0101 & 0.0104 \\
	Konjunktiv II & 0.0531 & 0.0281 \\
	Infinitive & 0.4407 & 0.3069 \\
}\mydataInf

\begin{figure*}[!ht]
\begin{floatrow}
\ffigbox{\pgfplotstableread[row sep=\\,col sep=&]{
	Tense & Präsens & Perfekt & Präteritum & Pluperfekt & Futur I & Futur II & Konjunktiv I & Konjunktiv II & - \\
	conditional1 & 0.0952110112615 & 0.00432260266181 & 0.224320327608 & 0.00699579115004 & 0.00722329655329 & 0.000455010806507 & 0.0210442498009 & 0.628995563645 & 0.0114321465135 \\
	conditional1Pr & 0 & 0 & 0.2 & 0 & 0 & 0 & 0 & 0.8 & 0 \\
	conditional2 & 0.0106809078772 & 0.0253671562083 & 0.0320427236315 & 0.0100133511348 & 0.000667556742323 & 0.000667556742323 & 0.00133511348465 & 0.917222963952 & 0.00200267022697 \\
	conditional2Pr & 0 & 0.2 & 0 & 0 & 0 & 0 & 0 & 0.8 & 0 \\
}\mydataCond
	
	\begin{tikzpicture}
    \tiny
	\begin{axis}[
	ybar stacked,
	bar width=0.5cm,
	width=0.5\textwidth,
	height=.45\textwidth,
	legend style={at={(0.844,1)},
		anchor=north,legend columns=-1,font=\tiny},
	symbolic x coords={conditional1, conditional1Pr, conditional2, conditional2Pr},
	xtick=data,legend columns=1, legend cell align=left,
	ymin=0,ymax=1,
	xticklabel style={rotate=45}
	]
	\addplot[imsblue1,fill=imsblue1!40!white,postaction={pattern=north west lines}] table[x=Tense,y=Präsens]{\mydataCond};
	\addplot[imsred1,fill=imsred1!40!white,postaction={pattern=vertical lines}] table[x=Tense,y=Perfekt]{\mydataCond};
	\addplot[Brown,fill=Brown!40!white,postaction={pattern=horizontal lines}] table[x=Tense,y=Präteritum]{\mydataCond};
	\addplot[OliveGreen,fill=OliveGreen!40!white,postaction={pattern=north east lines}] table[x=Tense,y=Pluperfekt]{\mydataCond};
	\addplot[RedViolet,fill=RedViolet!40!white,postaction={pattern=crosshatch dots}] table[x=Tense,y=Futur I]{\mydataCond};
	\addplot[Gray,fill=Gray!40!white,postaction={pattern=crosshatch}] table[x=Tense,y=Futur II]{\mydataCond};
	\addplot[BlueViolet,fill=BlueViolet!60!white,postaction={pattern=grid}] table[x=Tense,y=Konjunktiv I]{\mydataCond};
	\addplot[GreenYellow,fill=GreenYellow!40!white,postaction={pattern=fivepointed stars}] table[x=Tense,y=Konjunktiv II]{\mydataCond};
	\legend{Präsens, Perfekt, Präteritum, Pluperfekt, Futur I, Futur II, Konjunktiv I, Konjunktiv II}
	\end{axis}
	\end{tikzpicture}
    }{    \caption{				        German correspondences of the English conditionals in the News corpus.}
	\label{fig:condDistribNews}    }
    \ffigbox{    \begin{tikzpicture}
  \tiny
	\begin{axis}[
	ybar=.1cm,ymode = log,log origin=infty,ymajorgrids,
	bar width=.2cm,
	width=0.5\textwidth,
	height=.45\textwidth,
	legend style={at={(0.793,1)},
		anchor=north,legend columns=-1},
	symbolic x coords={Präsens, Präteritum, Perfekt, Pluperfekt,Futur I, Futur II, Konjunktiv I, Konjunktiv II, Infinitive},
	xtick=data,
	ytick={0.001,0.01,0.05,0.1,0.3,0.6},
	yticklabels={0.001,0.01,0.05,0.1,0.3,0.6},
	ymin=0,ymax=0.6,
	xticklabel style={rotate=40}
	]
	\addplot[imsred1,fill=imsred1!40!white,postaction={pattern=north west lines}] table[x=Tense,y=News]{\mydataInf};
	\addplot[imsblue1,fill=imsblue1!40!white,postaction={pattern=north east lines}] table[x=Tense,y=Europarl]{\mydataInf};
	\legend{News, Europarl}
	\end{axis}
	\end{tikzpicture}
}{  \caption{		        German correspondences of the English non-finite VCs -- gerunds and to-infinitives.}
	\label{fig:nonfinNewsEP}
}
\end{floatrow}
\end{figure*}

\paragraph*{Tense interchangeability}
\label{sec:tenseInterchange}
In the data from the News corpus, we identified 190 occurrences of the auxiliary \emph{sein (to be)} in one of the composed past tenses (i.e., \emph{Perfekt} and \emph{Plusquamperfekt}) in active voice in contrast to 10,247 occurrences in the simple past tense \emph{Präteritum}. This lexical preference is also given for a few additional full verbs in German as shown by counts derived from the Crawl corpus: \emph{denken} (819 vs.\ 354), \emph{stehen} (3083 vs.\ 98), \emph{geben} (7220 vs.\ 1523), \emph{ziehen} (1565 vs.\ 145). The data suggests the same preference also for the passive voice: \emph{denken} (184 vs.\ 38), \emph{geben} (1517 vs.\ 395), \emph{ziehen} (380 vs.\ 78).

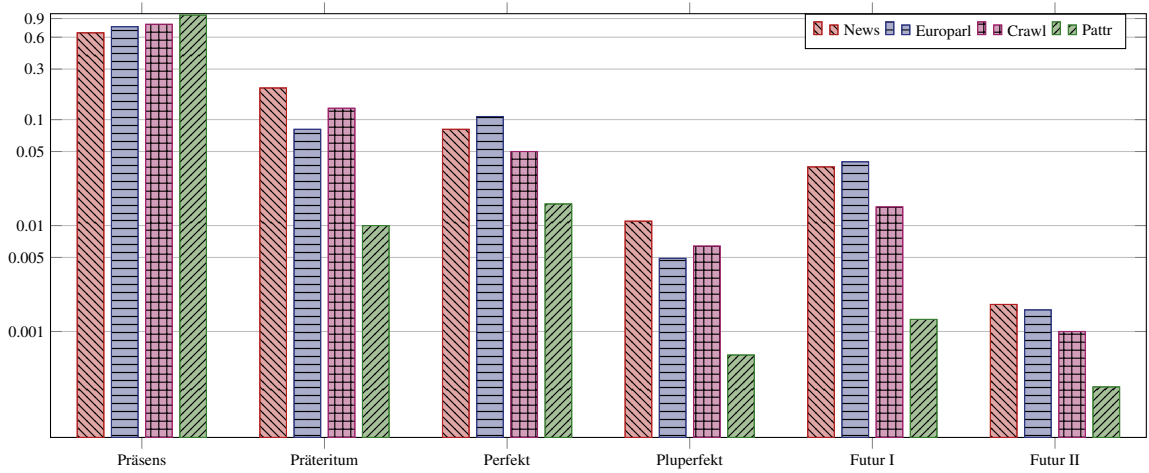
\begin{figure*}[!ht]
	\centering
	\begin{tikzpicture}
    \tiny
	\begin{axis}[
	ybar=.1cm,ymode = log,log origin=infty,
	bar width=.35cm,ymajorgrids,
	width=\textwidth,
	height=.45\textwidth,
	legend style={at={(0.835,1)},
		anchor=north,legend columns=-1},
	symbolic x coords={Präsens, Präteritum, Perfekt, Pluperfekt,Futur I, Futur II},
	xtick=data,
	ytick={0.001,0.005,0.01,0.05,0.1,0.3,0.6,0.9},
	yticklabels={0.001,0.005,0.01,0.05,0.1,0.3,0.6,0.9},
	ymin=0.0001,ymax=1,
	y label style={yshift=-0.3em},xticklabel style={rotate=0},
	]
	\addplot[imsred1,fill=imsred1!40!white,postaction={pattern=north west lines}] table[x=Tense,y=News]{\mydataIndDE};
	\addplot[imsblue1,fill=imsblue1!40!white,postaction={pattern=horizontal lines}] table[x=Tense,y=Europarl]{\mydataIndDE};
	\addplot[RedViolet,fill=RedViolet!40!white,postaction={pattern=grid}] table[x=Tense,y=Crawl]{\mydataIndDE};
	\addplot[OliveGreen,fill=OliveGreen!40!white,postaction={pattern=north east lines}] table[x=Tense,y=Pattr]{\mydataIndDE};
	\legend{News, Europarl, Crawl, Pattr} 
	\end{axis}
	\end{tikzpicture}
	\caption{Relative frequencies of the indicative active tense 
		forms in four different German corpora.}	\label{fig:tenseDistrCompDE}
\end{figure*}

\paragraph*{Contextual specifics}
\label{sec:domainReg}
Direct comparison of the tense frequencies extracted from the German data (Figure \ref{fig:tenseDistrCompDE}) shows variation in the usage of tenses in different domains (or registers). Being the most frequent tense form in all corpora,  \emph{Präsens}, however, differs in its relative frequency. For instance, in News, the relative frequency of \emph{Präsens} is 9\% lower than in Europarl (0.66 vs. 0.75), while \emph{Präsens} represents 97\% of the tense forms in Pattr (medical texts). 
We also observe variation in the past tense use within the respective corpora. For instance, in Europarl, \emph{Präteritum} and \emph{Perfekt} have almost equal relative frequency (0.08 and 0.10, respectively), whereas News clearly prefers the narrative tense \emph{Präteritum} over the discussing tense \emph{Perfekt} (0.19 vs. 0.08, respectively). 

\pgfplotstableread[row sep=\\,col sep=&]{
	Tense & News & Europarl & Crawl & Pattr \\
	Präsens & 0.660 & 0.754 & 0.794 & 0.970 \\
	Präteritum & 0.199 & 0.081 & 0.128 &  0.010 \\
	Perfekt & 0.081 & 0.107 & 0.050 & 0.016 \\
	Pluperfekt &0.011 & 0.0049 & 0.0064 & 0.0006 \\
	Futur I & 0.0358 & 0.040 & 0.015 & 0.0013 \\
	Futur II & 0.0018 & 0.0016 & 0.001 & 0.0003 \\
}\mydataIndDE

\subsection{Modeling tense and mood}
\label{sec:autoTM}

\paragraph*{Many-to-many relation}
\label{sec:manyMany}

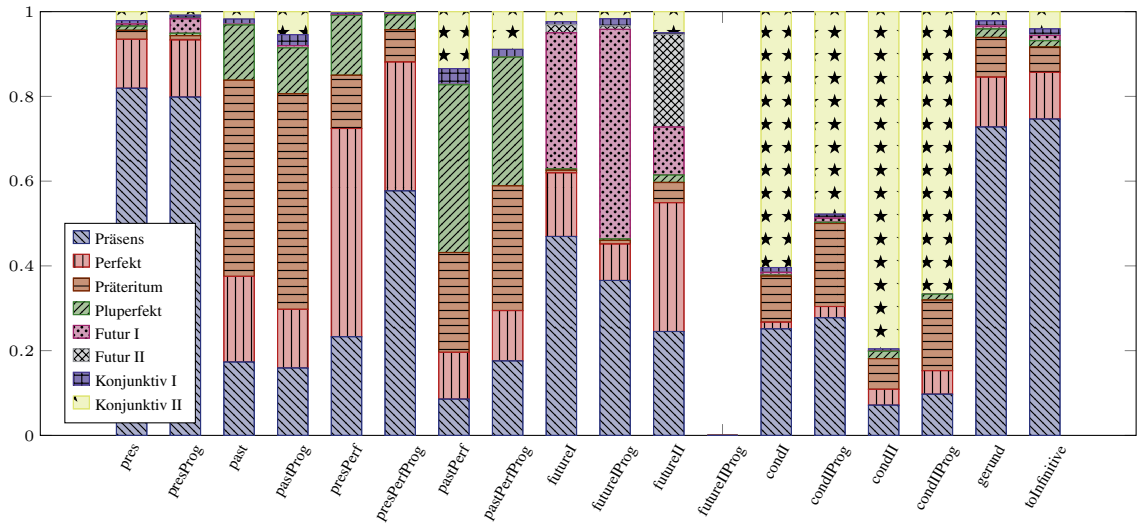
\begin{figure*}[!ht]
	\centering
	\begin{tikzpicture}
    \tiny
	\begin{axis}[
	ybar stacked,
	bar width=.4cm,
	width=\textwidth,
	height=.45\textwidth,
	legend style={at={(0.142,0.265)},
		anchor=east,legend columns=-1,font=\tiny},
	symbolic x coords={pres, presProg, past, pastProg, presPerf, presPerfProg, pastPerf, pastPerfProg, futureI, futureIProg, futureII, futureIIProg, condI, condIProg, condII, condIIProg, gerund,  toInfinitive},
	xtick=data,legend columns=1, legend cell align=left,
	ymin=0,ymax=1,
	xticklabel style={rotate=60}
    ]
	\addplot[imsblue1,fill=imsblue1!40!white,postaction={pattern=north west lines}] table[x=Tense,y=Präsens]{\mydataAll};
	\addplot[imsred1,fill=imsred1!40!white,postaction={pattern=vertical lines}] table[x=Tense,y=Perfekt]{\mydataAll};
	\addplot[Brown,fill=Brown!40!white,postaction={pattern=horizontal lines}] table[x=Tense,y=Präteritum]{\mydataAll};
	\addplot[OliveGreen,fill=OliveGreen!40!white,postaction={pattern=north east lines}] table[x=Tense,y=Pluperfekt]{\mydataAll};
	\addplot[RedViolet,fill=RedViolet!40!white,postaction={pattern=crosshatch dots}] table[x=Tense,y=Futur I]{\mydataAll};
	\addplot[Gray,fill=Gray!40!white,postaction={pattern=crosshatch}] table[x=Tense,y=Futur II]{\mydataAll};
	\addplot[BlueViolet,fill=BlueViolet!60!white,postaction={pattern=grid}] table[x=Tense,y=Konjunktiv I]{\mydataAll};
	\addplot[GreenYellow,fill=GreenYellow!40!white,postaction={pattern=fivepointed stars}] table[x=Tense,y=Konjunktiv II]{\mydataAll};
	\legend{Präsens, Perfekt, Präteritum, Pluperfekt, Futur I, Futur II, Konjunktiv I, Konjunktiv II}
	\end{axis}
	\end{tikzpicture}
	\caption{Distribution of tense translations derived from the News, Europarl and Crawl corpus.}
	\label{fig:overallTenseDistribNewsEPCrawl}
\end{figure*}

Corpus analyses of human translations presented in Section \ref{sec:tmHT} show that the respective monolingual, as well as bilingual linguistic
tense-related specifics in English and German result in a many-to-many relation.  
Figure \ref{fig:overallTenseDistribNewsEPCrawl} illustrates this relation on the basis of distributions of tense transformation patterns derived from our data. A formal description of the respective many-to-many relation  requires knowledge on different linguistic levels: lexical, syntactic and semantic/pragmatic.

One of the reasons for the many-to-many relation is the different granularity of the tense systems in the two languages. While there are tenses in English which do not have a direct counterpart in German, some tense forms in German do not have a direct counterpart in English (\emph{Konjunktiv I}) either. 
\pgfplotstableread[row sep=\\,col sep=&]{
	Tense & Präsens & Perfekt & Präteritum & Pluperfekt & Futur I & Futur II & Konjunktiv I & Konjunktiv II & - \\
	pres & 0.81918135004 & 0.115369795698 & 0.0214016747603 & 0.0101149776751 & 0.00488866344449 & 0.000311653842023 & 0.00762391335095 & 0.0211079711892 & 0 \\
	presProg & 0.798405730129 & 0.13488331793 & 0.0100450554529 & 0.00573590573013 & 0.0340168669131 & 0.0012188077634 & 0.00741104436229 & 0.00828327171904 & 0 \\
	presPerf & 0.232838227101 & 0.491376142439 & 0.125730508897 & 0.140980808213 & 0.00144027311846 & 0.000160030346495 & 0.00429711115589 & 0.00317689873043 & 0 \\
	presPerfProg & 0.577198697068 & 0.304017372421 & 0.0762214983713 & 0.0349619978284 & 0.00217155266015 & 0 & 0.00304017372421 & 0.00238870792617 & 0 \\
	past & 0.17318645162 & 0.202017888194 & 0.463010589228 & 0.130442554048 & 0.00156703907706 & 9.71807179573e-05 & 0.012225681393 & 0.0174526157231 & 0 \\
	pastProg & 0.159038461538 & 0.138365384615 & 0.508846153846 & 0.108365384615 & 0.00528846153846 & 0 & 0.0258653846154 & 0.0542307692308 & 0 \\
	pastPerf & 0.0855924117416 & 0.110344053432 & 0.23567379469 & 0.395857888533 & 0.000785766402874 & 0.000224504686535 & 0.03709939945 & 0.134422181063 & 0 \\
	pastPerfProg & 0.175799086758 & 0.118721461187 & 0.294520547945 & 0.303652968037 & 0 & 0 & 0.0182648401826 & 0.0890410958904 & 0 \\
	futureI & 0.469405439534 & 0.150226620408 & 0.00623649055668 & 0.0030151627898 & 0.321275903502 & 0.0188673167306 & 0.00742516050266 & 0.0235479059759 & 0 \\
	futureIProg & 0.365558912387 & 0.0861027190332 & 0.00848245410179 & 0.00348594004183 & 0.494538693934 & 0.00801766209621 & 0.017081106205 & 0.0167325122008 & 0 \\
	futureII & 0.245148771022 & 0.304010349288 & 0.0478654592497 & 0.0174644243208 & 0.11319534282 & 0.218628719276 & 0.00388098318241 & 0.0498059508409 & 0 \\
	futureIIProg & 0 & 0 & 0 & 0 & 0 & 0 & 0 & 0 & 0 \\
	condI & 0.251467555132 & 0.0157140178559 & 0.108851484863 & 0.00220674137856 & 0.00630291491786 & 0.000198318261145 & 0.0114087086957 & 0.603850258895 & 0 \\
	condIProg & 0.277693856999 & 0.0264350453172 & 0.195871097684 & 0.00302114803625 & 0.00981873111782 & 0.000503524672709 & 0.00881168177241 & 0.477844914401 & 0 \\
	condII & 0.0713068375486 & 0.0373576930943 & 0.0724657440862 & 0.0175199400095 & 0.000818051673597 & 0.000340854863999 & 0.00443111323199 & 0.795759765492 & 0 \\
	condIIProg & 0.0972222222222 & 0.0555555555556 & 0.166666666667 & 0.0138888888889 & 0 & 0 & 0 & 0.666666666667 & 0 \\
	gerund & 0.727848316133 & 0.117599130907 & 0.0937160510592 & 0.0200638240087 & 0.00662004345464 & 0.00039041281912 & 0.0122895165671 & 0.0214727050516 & 0 \\
	toInfinitive & 0.746543171942 & 0.110096620478 & 0.0599281077311 & 0.0149539821196 & 0.0122177838257 & 0.00071697174546 & 0.0151149349604 & 0.0404284271981 & 0 \\
}\mydataAll
\paragraph*{Tense/mood-related contextual features}
\label{sec:tmCtxFeat}

\begin{table*}[!ht]
	\centering
	\tiny
	\begin{tabular}{|m{3cm}||m{5.5cm}|m{6cm}|}
		\hline
		\multicolumn{1}{|c||}{\bf Textual property} & \multicolumn{1}{c|}{\bf Lexical/syntactic level} & 
		\multicolumn{1}{c|}{\bf Annotation tool availability} \\
		\hline \hline
		\multirow{10}{*}{\bf Tense} & VC, main verb & POS tagging and  parse trees\\
		& tense, mood, voice &  TMVannotator \cite{ramm17:acl} \\
		& temporal expressions (NPs and PPs): &  TARSQI \cite{verhagen05:acl} \\
		& \hspace{1cm} head, preposition, adjective, adverb & 
				POS tagging + parse trees \\
								& temporal ordering & TARSQI \cite{verhagen05:acl} \\
		\hline
		\multirow{1}{*}{\bf Aspect} & auxiliary (combination) & POS tagging and parse trees + mapping rules \\ 
		\hline
		\multirow{7}{*}{\bf Aktionsart} & event/state/progress &  sitent \cite{friedrich16:acl} \\
		& subject NP: & parse trees \\
		& \hspace{1cm} determiner, quantifier & semantic properties \\
		& \hspace{1cm} number & POS tagging \\ 
		& \hspace{1cm} mass, count & WordNet\\
		\hline
		{\bf Domain/ genre} & & - \\
		\hline
		{\bf Reported speech} & &  QSample \cite{scheible16:acl} \\
		\hline
		{\bf Conditional clauses} & & - \\
		\hline
	\end{tabular}
	\caption{\label{tab:TAtoLexSynt}Mapping of the different textual properties to the corresponding lexical/syntactic levels. Column \emph{Tool availability} lists tools for automatic annotation of the English texts with the respective information.}
\end{table*}

For automatic modeling of tense and mood, textual characteristics discussed in the preceding sections need to be mapped to the specific contextual information overtly given in a sentence. The respective contextual features are summarized in Table \ref{tab:TAtoLexSynt}. Many of these features can be derived from parsed and POS-tagged data. However, some of them require access to other annotation tools, as well as lexical databases which include information about semantic properties of the words/ Automatic annotation of the temporal ordering can be done with the tool {\textsc TARSQI} \cite{verhagen05:acl}. Information about tense, mood and voice of the VCs in the English texts can be obtained with the {\textsc TMV annotator} \cite{ramm17:acl}. Information about Aktionsart in terms of state, event and progress can be gained from the output of the tool {\textsc sitent} \cite{friedrich16:acl}. Currently, no tools are publicly available for automatic identification of conditionals in English, which is important for translation of the German subjunctive mood. However, the set of syntactic rules described
by
\newcite{olivas05:icai} can be re-used to identify the respective contexts in English. To our knowledge, there are no publicly available tools for automatic annotation of texts with genre and/or domain information, although there has been ongoing research in this area \cite{santini07:phd,sharoff10:lrec,petrenz14,biber16:jrds}. 
As seen from Table \ref{tab:TAtoLexSynt}, textual properties carrying pragmatic information represent their own subtasks in NLP. Tools for annotation of the respective information are mostly based on classification models that use many different subtask-related information. While predicted annotations are correct in many cases, they may also be erroneous, consequently having negative impact on training a tense/mood translation model. Instead of using outputs of many different tools requiring a complex processing pipeline, one might train a model directly with the features used to train models for predicting each of the relevant textual properties. 

The information on the distribution of various tenses and mood in the bilingual data is also important. Therefore, training data  should be carefully preselected to account for this specific distribution.

\section{Discussion and Conclusion}
\label{sec:conclusion}
The
paper describes a contrastive analysis of English and German tense and mood by means of parallel data.
We provide an overview of the categories available in both language systems, point to the existing asymmetries providing corpus-based evidence from human translations and formulate assumptions on their impact on MT. Our translation data shows considerable amount of variation of tense/mood translation between the two languages. Translations also
vary
a lot leading to many unexpected tense/mood correspondences. The observed variation may be explained by a number of different factors which are not only related to the differences on lexical/syntactic level between the considered languages, but also to a number of pragmatic factors, including the process of translation. 
We also show that modeling tense/mood for MT requires additional information beyond the morpho-syntactic properties of the source, 
and we discuss tools for obtaining this information,
which can (and should) be used in future modeling research.
Beyond directly improving the modeling, an interesting future consideration would be to give the
translation system
user control of document-level tense and mood choices (e.g., by introducing a parameter for how choices for tense and mood in indirect speech should be made).

\bibliography{colingbib}

\begin{thebibliography}{29}
\expandafter\ifx\csname natexlab\endcsname\relax\def\natexlab#1{#1}\fi

\bibitem[{Baker(1993)}]{baker93}
Mona Baker. 1993.
\newblock Corpus linguistics and translation studies: Implications and
  applications.
\newblock In G.~Francis Baker~M. and E.~Tognini-Bonelli, editors, \emph{Text
  and Technology: in Honour of John Sinclair}, pages 233--250. Benjamins,
  Amsterdam.

\bibitem[{Biber(1995)}]{biber1995}
Douglas Biber. 1995.
\newblock \emph{Dimensions of Register Variation: A Cross-Linguistic
  Comparison}.
\newblock Cambridge University Press.

\bibitem[{Biber and Egbert(2016)}]{biber16:jrds}
Douglas Biber and Jesse Egbert. 2016.
\newblock Using grammatical features for automatic register identification in
  an unrestricted corpus of documents from the open web.
\newblock \emph{Journal of Research Design and Statistics in Linguistics and
  Communication Science}, 2.

\bibitem[{Bohnet and Nivre(2012)}]{bohnet12:emnlp}
Bernd Bohnet and Joakim Nivre. 2012.
\newblock {A Transition-Based System for Joint Part-of-Speech Tagging and
  Labeled Non-Projective Dependency Parsing}.
\newblock In \emph{In Proceedings of EMNLP-CoNLL}, Jeju, Korea.

\bibitem[{Bojar et~al.(2015)Bojar, Chatterjee, Federmann, Haddow, Huck, Hokamp,
  Koehn, Logacheva, Monz, Negri, Post, Scarton, Specia, and Turchi}]{wmt15}
Ond\v{r}ej Bojar, Rajen Chatterjee, Christian Federmann, Barry Haddow, Matthias
  Huck, Chris Hokamp, Philipp Koehn, Varvara Logacheva, Christof Monz, Matteo
  Negri, Matt Post, Carolina Scarton, Lucia Specia, and Marco Turchi. 2015.
\newblock Findings of the 2015 workshop on statistical machine translation.
\newblock In \emph{Proceedings of the Tenth Workshop on Statistical Machine
  Translation}, pages 1--46, Lisbon, Portugal. Association for Computational
  Linguistics.

\bibitem[{Copeland et~al.(1991)Copeland, Durand, Krauwer, and
  Maegaar}]{copeland-et-al:1991}
Charles Copeland, Jacques Durand, Steven Krauwer, and Bente Maegaar. 1991.
\newblock The {E}urotra linguistic specifications.
\newblock Technical report, Office for Official Publications of the Commission
  of the European Communities, Brussels/Luxembourg.
\newblock Studies in Machine Translation and Natural Language Processing 1.

\bibitem[{Csipak(2015)}]{csipak15:gram}
Eva Csipak. 2015.
\newblock \emph{{Free factive subjunctives in German}}.
\newblock Doctoral thesis. Nieders\"{a}chsische Staats- und
  Universit\"{a}tsbibliothek G\"{o}ttingen.

\bibitem[{Friedrich and Palmer(2016)}]{friedrich16:acl}
Annemarie Friedrich and Alexis Palmer. 2016.
\newblock Situation entity types: automatic classification of clause-level
  aspect.
\newblock In \emph{Proceedings of ACL}, Berlin, Germany.

\bibitem[{Gellerstam(1986)}]{gellerstam86}
Martin Gellerstam. 1986.
\newblock {T}ranslationese in {S}wedish novels translated from {E}nglish.
\newblock In L.~Wollin and H.~Lindquist, editors, \emph{Translation Studies in
  Scandinavia}, pages 88--95. CWK Gleerup, Lund.

\bibitem[{Gispert and {Mari\~no}(2008)}]{degispertmarino:08}
{Adri\`a de} Gispert and Jose~B. {Mari\~no}. 2008.
\newblock {On the impact of morphology in English to Spanish statistical MT}.
\newblock \emph{Speech Communication}, 50(11-12):1034--1046.

\bibitem[{Hawkins(2015)}]{hawkins2015comparative}
J.A. Hawkins. 2015.
\newblock \emph{A Comparative Typology of English and German: Unifying the
  Contrasts}.
\newblock Routledge Library Editions: The English Language. Taylor \& Francis.

\bibitem[{Koehn et~al.(2007)Koehn, Hoang, Birch, Callison-Burch, Federico,
  Bertoldi, Cowan, Shen, Moran, Zens, Dyer, Bojar, Constantin, and
  Herbst}]{koehn07:acl}
Philipp Koehn, Hieu Hoang, Alexandra Birch, Chris Callison-Burch, Marcello
  Federico, Nicola Bertoldi, Brooke Cowan, Wade Shen, Christine Moran, Richard
  Zens, Chris Dyer, Ondrej Bojar, Alexandra Constantin, and Evan Herbst. 2007.
\newblock Moses: Open source toolkit for statistical machine translation.
\newblock In \emph{Proceedings of ACL, demonstration session}, Prague, Czech
  Republic.

\bibitem[{K\"onig and Gast(2012)}]{koenig12:endeconstr}
Ekkehard K\"onig and Volker Gast. 2012.
\newblock \emph{Understanding English-German constrasts}.
\newblock Number~29 in Grundlagen der Anglistik und Amerikanistik. Erich
  Schmidt Verlag.

\bibitem[{Lo\'{a}iciga et~al.(2014)Lo\'{a}iciga, Meyer, and
  Popescu-Belis}]{loaiciga14:lrec}
Sharid Lo\'{a}iciga, Thomas Meyer, and Andrei Popescu-Belis. 2014.
\newblock {English-French Verb Phrase Alignment in Europarl for Tense
  Translation Modeling}.
\newblock In \emph{Proceedings of the The 9th Language Resources and Evaluation
  Conference (LREC)}, Reykjavik, Iceland.

\bibitem[{Neumann(2013)}]{neumann13:regVar}
Stella Neumann. 2013.
\newblock \emph{{Contrastive register variation. A quantitative approach to the
  comparison of English and German}}.
\newblock Trends in Linguistics. Studies and Monographs. De Gruyter Mouton.

\bibitem[{Och and Ney(2003)}]{och03:asc}
{Franz Josef} Och and Hermann Ney. 2003.
\newblock {A Systematic Comparison of Various Statistical Alignment Models}.
\newblock \emph{Computational Linguistics}, 29(1):19--51.

\bibitem[{Olivas et~al.(2005)Olivas, Puente, and Tejado}]{olivas05:icai}
José~A. Olivas, Cristina Puente, and Andrea Tejado. 2005.
\newblock {Searching for causal relations in text documents for ontological
  application}.
\newblock In \emph{Proceedings of ICAI}, Las Vegas, Nevada, USA.

\bibitem[{Petrenz(2014)}]{petrenz14}
Philipp Petrenz. 2014.
\newblock \emph{{Cross-Lingual Genre Classification}}.
\newblock Ph.D. thesis, School of Informatics, University of Edinburgh,
  Scotland.

\bibitem[{Ramm and Fraser(2016)}]{ramm16:wmt}
Anita Ramm and Alexander~M. Fraser. 2016.
\newblock {Modeling verbal inflection for English to German SMT}.
\newblock In \emph{Proceedings of WMT}, Berlin, Germany.

\bibitem[{Ramm et~al.(2017)Ramm, Lo{\'a}iciga, Friedrich, and
  Fraser}]{ramm17:acl}
Anita Ramm, Sharid Lo{\'a}iciga, Annemarie Friedrich, and Alexander Fraser.
  2017.
\newblock {Annotating tense, mood and voice for English, French and German}.
\newblock In \emph{Proceedings of ACL, demonstration session}, Vancouver,
  Canada.

\bibitem[{Sammon(2002)}]{sammon02:engram}
Geoff Sammon. 2002.
\newblock \emph{{Exploring English grammar}}.
\newblock Cornelson Verlag.

\bibitem[{Santini(2007)}]{santini07:phd}
Marina Santini. 2007.
\newblock \emph{{Automatic Identification of Genre in Web Pages}}.
\newblock Doctoral thesis. University of Brighton.

\bibitem[{Scheible et~al.(2016)Scheible, Klinger, and
  Pad\'{o}}]{scheible16:acl}
Christian Scheible, Roman Klinger, and Sebastian Pad\'{o}. 2016.
\newblock {Model Architectures for Quotation Detection}.
\newblock In \emph{Proceedings of ACL}, Berlin, Germany.

\bibitem[{Sharoff et~al.(2010)Sharoff, Wu, and Markert}]{sharoff10:lrec}
Serge Sharoff, Zhili Wu, and Katja Markert. 2010.
\newblock {The Web Library of Babel: evaluating genre collections}.
\newblock In \emph{Proceedings of LREC}, Malta.

\bibitem[{Tajiri et~al.(2012)Tajiri, Komachi, and Matsumoto}]{tajiri12:acl}
Toshikazu Tajiri, Mamoru Komachi, and Yuji Matsumoto. 2012.
\newblock {Tense and aspect error correction for ESL learners using global
  context}.
\newblock In \emph{Proceedings of the 50th Annual Meeting of the Association
  for Computational Linguistics (ACL): Short Papers - Volume 2}, Jeju Island,
  Korea.

\bibitem[{Teich(2003)}]{teich03}
Elke Teich. 2003.
\newblock \emph{Cross-Linguistic Variation in System und Text. A Methodology
  for the Investigation of Translations and Comparable Texts}.
\newblock Mouton de Gruyter, Berlin.

\bibitem[{Verhagen et~al.(2005)Verhagen, Mani, Saurí, ~, Littman, and
  Pustejovsky}]{verhagen05:acl}
Marc Verhagen, Inderjeet Mani, Roser Saurí, Robert ~, Knippen, Jess Littman,
  and James Pustejovsky. 2005.
\newblock {Automating Temporal Annotation with TARSQI}.
\newblock In \emph{Proceedings of ACL, demonstration session}, Ann Arbor,
  Michigan, USA.

\bibitem[{Weinrich(2001)}]{weinrich64:tempus}
Harald Weinrich. 2001.
\newblock \emph{{Tempus. Besprochene und erzählte Welt}}, 6 edition.
\newblock C.H.Beck.

\bibitem[{Ye et~al.(2006)Ye, Fossum, Victoria, and Abney}]{ye06:sighan}
Yang Ye, Li~Fossum, Victoria, and Steven Abney. 2006.
\newblock {Latent Features in Automatic Tense Translation between Chinese and
  English}.
\newblock In \emph{Proceedings of the Seventh SIGHAN Workshop on Chinese
  Language Processing}, Sidney, Australia.

\end{thebibliography}
\bibliographystyle{acl_natbib_nourl}

\end{document}